\documentclass[sigconf]{acmart}
\AtBeginDocument{%
}

\copyrightyear{2026}
\acmYear{2026}
\setcopyright{cc}
\setcctype{by}
\acmConference[MM '26]{Proceedings of the 34th ACM International Conference on Multimedia}{November 10--14, 2026}{Rio de Janeiro, Brazil}
\acmBooktitle{Proceedings of the 34th ACM International Conference on Multimedia (MM '26), November 10--14, 2026, Rio de Janeiro, Brazil}
\acmDOI{10.1145/3767308.3837679}
\acmISBN{979-8-4007-2213-4/2026/11}

\acmSubmissionID{}

\usepackage{overpic}

\usepackage{subcaption}

\usepackage{multirow}
\usepackage{multicol}

\usepackage{xcolor}
\PassOptionsToPackage{numbers}{natbib}
\definecolor{url}{RGB}{0,73,147}
\definecolor{mypink}{HTML}{bc4749}
\definecolor{mygray}{gray}{0.85}
\definecolor{rgbgray}{gray}{0.95}
\definecolor{todogreen}{RGB}{88,128,96}

\usepackage{array}
\usepackage{colortbl}

\usepackage{pifont}  
\usepackage{bbding}    

\usepackage{caption}
\usepackage{enumitem}
\usepackage{multirow}
\makeatletter
\newcommand{\thickhline}{%
\noalign {\ifnum 0=`}\fi \hrule height 1pt
\futurelet \reserved@a \@xhline
}
\makeatother

\usepackage{xspace}

\makeatletter
\DeclareRobustCommand\onedot{\futurelet\@let@token\@onedot}
\def\@onedot{\ifx\@let@token.\else.\null\fi\xspace}

\makeatother

\usepackage{hyperref}

\begin{document}

\title{MAC 2026: Advancing Micro-Action Analysis Towards Fine-Grained Understanding}


\author{Kun Li}
\orcid{0000-0001-5083-2145}
\affiliation{
\institution{
United Arab Emirates University
}
\city{Abu Dhabi}
\country{United Arab Emirates}
}
\email{kunli.hfut@gmail.com}

\author{Dan Guo}
\orcid{0000-0003-2594-254X}
\authornote{Corresponding author}
\affiliation{
\institution{
Hefei University of Technology}
\city{Hefei}
\country{China}
}
\email{guodan@hfut.edu.cn}

\author{Jihao Gu}
\orcid{0009-0009-0141-4807}
\affiliation{
\institution{
University College London
}
\city{London}
\country{United Kingdom}
}
\email{jihao.gu.23@ucl.ac.uk}

\author{Pengyu Liu}
\orcid{0000-0002-3396-3108}
\affiliation{
\institution{
Hefei University of Technology
}
\city{Hefei}
\country{China}
}
\email{lpynow@gmail.com}

\author{Xiaobai Li}
\orcid{0000-0003-4519-7823}
\affiliation{
\institution{Zhejiang University}
\institution{University of Oulu}
\city{Hangzhou}
\country{China}
}
\email{xiaobai.li@zju.edu.cn}

\author{Haoyu Chen}
\orcid{0000-0003-3267-2664}
\affiliation{
\institution{CMVS, University of Oulu}
\city{Oulu}
\country{Finland}
}
\email{chen.haoyu@oulu.fi}

\author{Yanbin Hao}
\orcid{0000-0002-0695-1566}
\affiliation{
\institution{
Hefei University of Technology
}
\city{Hefei}
\country{China}
}
\email{haoyanbin@hfut.edu.cn}

\author{Guoying Zhao}
\orcid{0000-0003-3694-206X}
\affiliation{
\institution{CMVS, University of Oulu}
\city{Oulu}
\country{Finland}
}
\email{guoying.zhao@oulu.fi}

\author{Meng Wang}
\orcid{0000-0002-3094-7735}
\affiliation{
\institution{Hefei University of Technology}
\city{Hefei}
\country{China}
}
\email{eric.mengwang@gmail.com}

\renewcommand{\shortauthors}{Kun Li et al.}

\begin{abstract}
Micro-Actions (MAs) are subtle and spontaneous human behaviors that provide important non-verbal cues in social interaction and affective communication. However, their short duration, weak motion patterns, and fine-grained semantic differences make them difficult to annotate, model, and evaluate in a standardized manner. To promote academic research on micro-action analysis, we proposed and have annually organized the Micro-Action Analysis Grand Challenge (MAC) as a public benchmark platform for this emerging field. The first two editions of MAC established standardized evaluation settings for micro-action recognition and detection, providing publicly accessible datasets and protocols. Building upon these editions, this paper presents the 3rd MAC, held in conjunction with ACM Multimedia 2026. Under the theme of moving from recognition to fine-grained micro-action understanding, this edition further expands the scope of the challenge beyond conventional recognition and detection. In particular, we introduce a new task named fine-grained micro-action understanding, evaluated with the assistance of multimodal large language models, aiming to assess models' ability to capture fine-grained semantic cues and interpret subtle human micro-actions at a deeper level. We summarize the datasets, task settings, evaluation protocols, competition results, and representative solutions from top-performing teams. Finally, we discuss future directions for micro-action analysis and its broader role in human-centric video understanding. 
\end{abstract}



\maketitle

\section{Introduction}
Human body actions~\cite{ekman1969repertoire,aviezer2012body,shiffrar2011seeing} are an important non-verbal communicative manner in social communication, in which people convey affective states, intentions, and interpersonal attitudes~\cite{chen2023smg,liu2021imigue,wang2026imigue}. 
Existing studies on human affective behavior have primarily focused on facial expressions~\cite{yan2014casme,zhao2023facial,li2017towards}, speech signals~\cite{zhao2025temporal}, and expressive body gestures~\cite{li2023data}. 
Different from these behavioral cues, Micro-Actions (MAs)~\cite{guo2024benchmarking,li2025mmad,li2025prototypical,chen2023smg,liu2021imigue,chen2024prototype} refer to subtle and spontaneous body movements that are closely related to inner psychological and emotional states~\cite{aviezer2012body}. 
Micro-Action analysis has attracted increasing attention in the fields of computer vision, affective computing, and human-centric video understanding~\cite{guo2024benchmarking,li2025mmad,gu2025mm,li2023joint,liu2024micro,liu2025online,chen20242nd,liu2026self,shen2026spatial}. 

Compared with conventional action recognition~\cite{li2025repetitive,li2023datae,wang2025exploiting}, micro-action analysis is more challenging due to the short duration, weak motion intensity, and high visual similarity of micro-actions. 
Despite its importance, research on micro-actions is still at an early stage due to a lack of publicly available benchmark datasets, standardized task definitions, and unified evaluation protocols. 

To bridge this gap, the MAC\footnote{Homepage: \href{https://sites.google.com/view/micro-action}{https://sites.google.com/view/micro-action}.} is designed as an annual series with a consistent yet evolving theme to drive continuous progress in the field~\cite{guo2024benchmarking,li2024advancing,wang2024instance,yu2024end,gong2024micro,guo2026rethinking,wang2025combatting,li2025progressive}. 
We have organized the 1st MAC~\cite{guo2024mac} and 2nd MAC~\cite{li2025mac} in conjunction with ACM MM 2024 and ACM MM 2025, respectively. 
Building upon this foundation, the 3rd edition of MAC, held with ACM MM 2026, further advances the theme from recognition to fine-grained micro-action understanding. 
In particular, we introduce a new Fine-grained Micro-Action Understanding task, evaluated with the assistance of multimodal large language models, to assess models' ability to capture subtle semantic cues and interpret micro-actions beyond predefined label prediction.

\section{Challenge Overview}\label{sec:overvview}

\subsection{Challenge Tracks}
\ding{113} \textbf{Track 1: Micro-Action Recognition (MAR).} MAR~\cite{guo2024benchmarking,guo2024mac,li2025prototypical,gu2025motion} aims to recognize spontaneous micro-actions characterized by subtle motion patterns and short temporal durations. 
Similar to conventional action recognition~\cite{wang2016temporal,I3D}, 
MAR is more challenging due to the extremely small motion amplitudes and high visual similarity among different action categories.

\noindent\ding{113} \textbf{Track 2: Multi-label Micro-Action Detection (MMAD).} 
Given that micro-actions may occur repeatedly over time and that multiple micro-actions can co-occur, we introduce the Multi-label Micro-Action Detection task~\cite{li2025mmad} to support a more fine-grained understanding of human bodily behavior. MMAD aims to identify and localize all micro-actions in a densely annotated video by determining their start and end times as well as their categories. 

\noindent\ding{114} \textbf{Track 3: Fine-grained Micro-Action Understanding (FMAU).} 
With the rapid development of multimodal large language models, recent MLLMs have shown strong potential in visual understanding and reasoning. Motivated by this, we propose the Fine-grained Micro-Action Understanding task~\cite{li2026ma} to examine whether MLLMs can capture, compare, and reason about subtle micro-actions. The task includes three levels: perceptual recognition, relational comprehension, and interpretive reasoning, gradually moving from recognizing visible actions to understanding relations and explaining the body movement.

\subsection{Competition Data}

\ding{113} \textbf{Track 1: Micro-Action Recognition.} 
In this track, we use the MA-52~\cite{guo2024benchmarking} dataset. It is a large-scale dataset designed for whole-body micro-action analysis, where professional interviewers were involved to elicit unconscious micro-behaviors from participants. 
The dataset includes 22,422 samples, covering 7 body-level classes and 52 fine-grained action classes.
The training and validation splits of MA-52 are adopted for model training and validation. The challenge final ranking is conducted on 1,138 samples drawn from the MA-52 test set. 

\noindent\ding{113} \textbf{Track 2: Multi-label Micro-Action Detection.} 
In this track, we use the MMA-52 dataset~\cite{li2025mmad}. 
It consists of 6,528 videos and 19,782 action instances across training, validation, and test splits. 
Note that this challenge only evaluates performance at the action-level for micro-action detection. The final rankings are based on detection performance on the test set.

\noindent \ding{114} \textbf{Track 3: Fine-grained Micro-Action Understanding.}
In this track, we adopt MA-Bench~\cite{li2026ma}, a fine-grained micro-action benchmark with 1,000 videos and 12,000 question-answer pairs. Following the ``Perception--Comprehension--Reasoning'' design, MA-Bench evaluates MLLMs from basic micro-action recognition to relation understanding and open-ended reasoning. It covers eight sub-tasks: CMAR and FMAR for coarse- and fine-grained recognition; SAD, MAD, MAS, and PPR for local motion, multi-part relations, temporal order, and proximity changes; and MADU and MARE for micro-action description and reasoning.

\subsection{Challenge Protocol}

\noindent\ding{113} \textbf{Development Stage} 
The challenge officially started on 4th May, 2026. Participants were allowed to register before the registration deadline on June 6, 2026. Each team was required to submit its member information and was assigned a unique authorized account for result submission to ensure a fair evaluation process.

\noindent\ding{113} \textbf{Testing Stage}
The testing data was released on 2nd June, 2026. During the testing period, from 2nd June to 9th June, 2026, all participating teams submitted their results through Kaggle using their assigned accounts.
The leaderboard was automatically updated according to the submitted results.

\subsection{Evaluation Metrics}

\ding{113} \textbf{Track 1: Micro-Action Recognition.}
Following the standard protocol in micro-action recognition~\cite{li2025prototypical,guo2024benchmarking,li2025mac}, we use the F1$_{mean}$ score to evaluate the performance. 
\begin{equation}
\rm F1_{\emph{mean}} =\frac{ \rm F1_{\emph{macro}}^{\emph{body}} \!+\! \rm F1_{\emph{micro}}^{\emph{body}} \!+\! \rm F1_{\emph{macro}}^{\emph{action}} \!+\! \rm F1_{\emph{micro}}^{\emph{action}}}{4},
\label{eq:map}
\end{equation}
where ``$body$'' and ``$action$'' denote the label granularity of body-level and action-level categories, respectively.  
The final results are ranked by the value of F1$_{mean}$. 

\noindent\noindent\ding{113} \textbf{Track 2: Multi-label Micro-Action Detection.}
We employ Detection-mAP~\cite{liu2024end,li2025mmad,guo2024mac} to evaluate the completeness of predicted instances. 
We report mAP values at tIoU thresholds from 0.1 to 0.9 with an interval of 0.1, together with their average mAP. 

\noindent\ding{114} \textbf{Track 3: Fine-grained Micro-Action Understanding.}
Following the benchmark evaluation protocol~\cite{li2026ma}, we report exact-match accuracy for closed-ended tasks, including CMAR, FMAR, SAD, MAD, MAS, and PPR. For open-ended tasks, including MADU and MARE, we use GPT-4o as the LLM judge to evaluate the quality of description and reasoning. The final ranking is determined by a weighted score combining the results from both task types.

\section{Proposed Solutions}~\label{sec:methods}

A total of 52 teams from 32 institutions and 6 countries/regions registered for the challenge. 
Here, we invited the leading teams in each track to share brief overviews of their approaches. The results are reported in Tab.~\ref{tab:all_results}. 

\subsection{Micro-Action Recognition}

\begin{figure}[t!]
\centering
\includegraphics[width=1\linewidth]{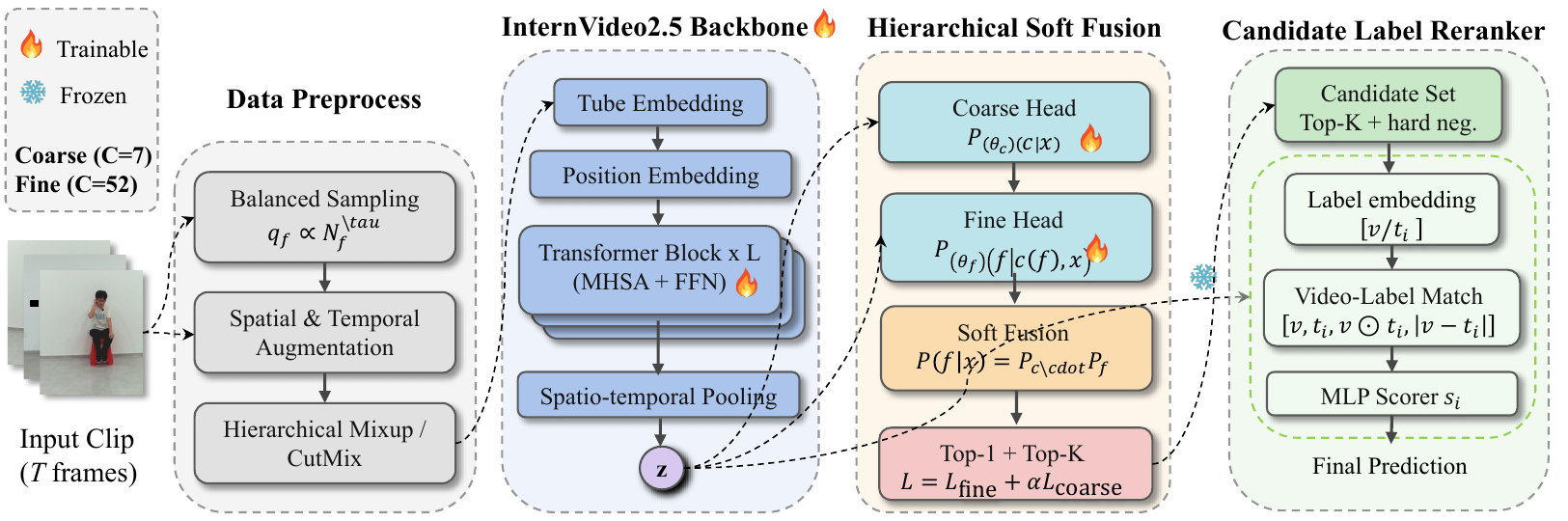}
\vspace{-1.5em}
\caption{Solution of team USTC-IAT-United2 for Track 1.}
\vspace{-1.5em}
\label{fig:track1_top1}
\end{figure}

\ding{182} Team ``USTC-IAT-United2'' is from the University of Science and Technology of China. 
As illustrated in Fig.~\ref{fig:track1_top1}, this team proposes a hierarchy-aware recognition framework built upon a fully fine-tuned InternVideo-2.5~\cite{wang2025internvideo2}. To mitigate the long-tailed distribution of MA-52~\cite{guo2024benchmarking}, they incorporate several imbalance-aware training strategies, including class-balanced sampling, loss reweighting, targeted data augmentation, and hierarchical Mixup. The framework jointly learns body-level and action-level representations through a hierarchical classification architecture, where coarse- and fine-grained predictions are fused according to the predefined label hierarchy. 
In addition, they introduce a lightweight candidate label reranking module to refine ambiguous predictions by re-ranking confusing action categories. These designs effectively exploit the hierarchical label structure and improve the discrimination of visually similar micro-actions, leading to superior recognition performance.

\begin{table}[t!]
\centering
\tabcolsep 6pt
\renewcommand{\arraystretch}{1.2}
\caption{The results of leading teams in MAC 2026.}
\vspace{-1.5em}
\resizebox{1.0\linewidth}{!}{
\begin{tabular}{|c|c|c|c|c|}
\hline
\thickhline
\rowcolor{mygray}
  & Rank & Team Name & Institutions & F1$_{mean}$  \\ \hline
\multirow{3}{*}{\rotatebox[origin=c]{90}{Track 1}} 
&  1st & USTC-IAT-United2~\cite{zhang2026softrerank} & USTC & 0.7999 \\ \cline{2-5}
&  2nd & HITWH & HIT (Weihai) \& USTC & 0.7504 \\ \cline{2-5}
&  3rd & salaheiyo1 & Aalto University & 0.7258 \\
\hline
\thickhline
\rowcolor{mygray}
  & Rank & Team Name& Institutions & avg. mAP \\ \hline
\multirow{3}{*}{\rotatebox[origin=c]{90}{Track 2}} 
&  1st & USTC-IAT-United1~\cite{zhu2026temporal} & USTC & 27.04 \\ \cline{2-5}
&  2nd & salaheiyo1 & Aalto University & 25.15 \\ \cline{2-5}
&  3rd & LR & FAU & 24.87 \\
\hline
\thickhline
\rowcolor{mygray}
  & Rank & Team Name& Institutions & Weighted Score \\ \hline
\multirow{2}{*}{\rotatebox[origin=c]{90}{Track 3}} 
&  1st & WHU-HUVPR~\cite{wang2026recognition} & Wuhan University & 57.16 \\ \cline{2-5}
&  2nd & GNC & Shanghai University & 46.29 \\
\hline
\thickhline
\end{tabular}
}
\vspace{-1.0em}
\label{tab:all_results}
\end{table}

\begin{figure}[t!]
\centering
\includegraphics[width=1\linewidth]{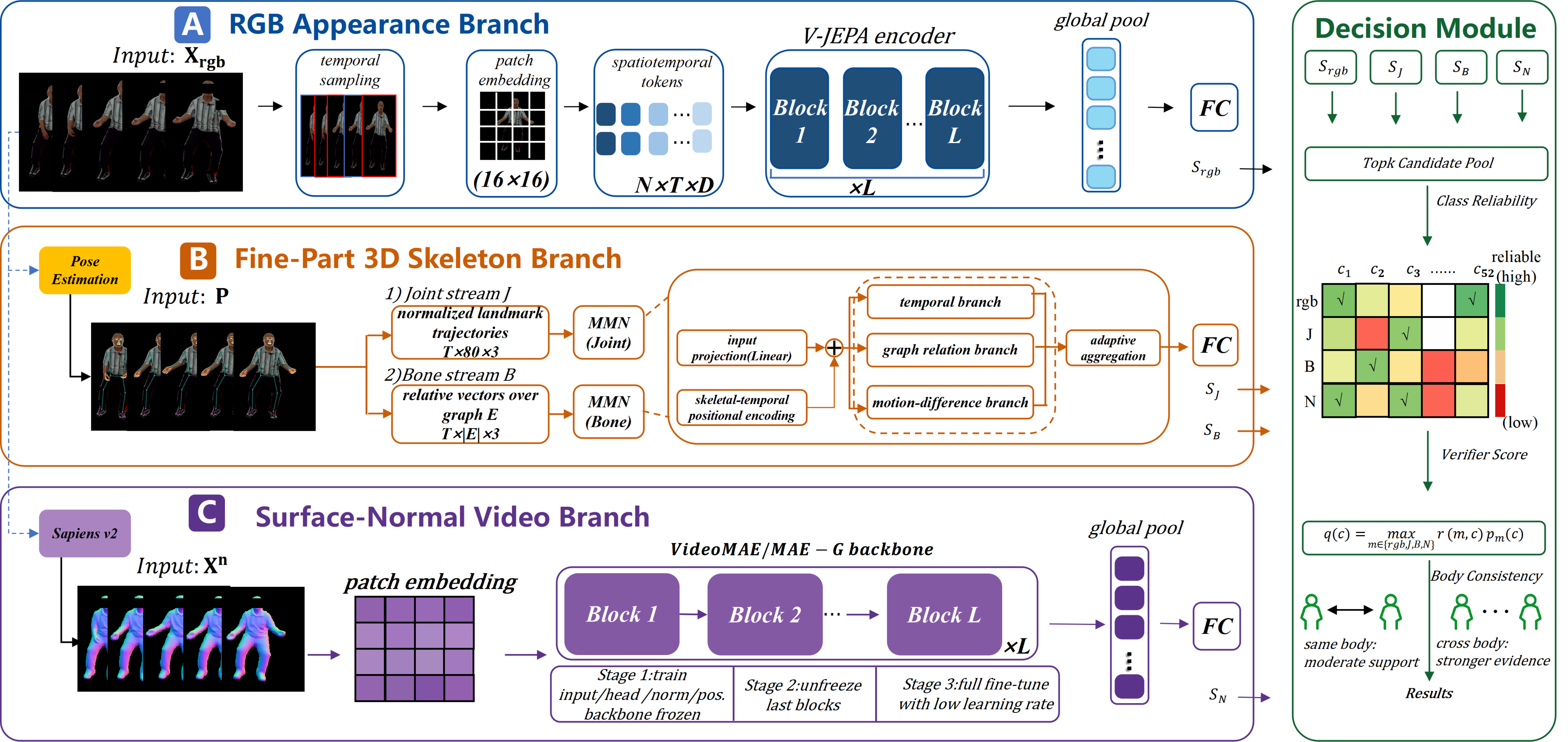}
\vspace{-1.5em}
\caption{Solution of team HITWH for Track 1.}
\vspace{-1.5em}
\label{fig:track1_top2}
\end{figure}

\ding{183} Team ``HITWH'' is from Harbin Institute of Technology (Weihai) and the University of Science and Technology of China.
As shown in Fig.~\ref{fig:track1_top2}, this team proposes GAF-MAR, a geometry-aware multi-representation framework for fine-grained micro-action recognition. 
Their solution adopts a three-branch framework, consisting of an RGB appearance branch~\cite{bardes2024revisiting}, a fine-part 3D skeleton branch~\cite{sarandi2024neural,gu2025motion}, and a surface-normal video branch~\cite{khirodkar2024sapiens,tong2022videomae,wang2023videomae}, to exploit complementary visual representations and capture subtle localized motion cues from different perspectives.
The framework independently learns representation-specific features and combines their predictions through a decision-level fusion strategy. To further improve recognition accuracy, the team introduces a class-aware expert verification and candidate reranking mechanism that exploits class-specific reliability and body-level consistency to refine ambiguous predictions. 
By effectively integrating appearance, skeletal motion, and geometric information, the proposed framework achieves competitive performance.

\begin{figure}[t!]
\centering
\includegraphics[width=1\linewidth]{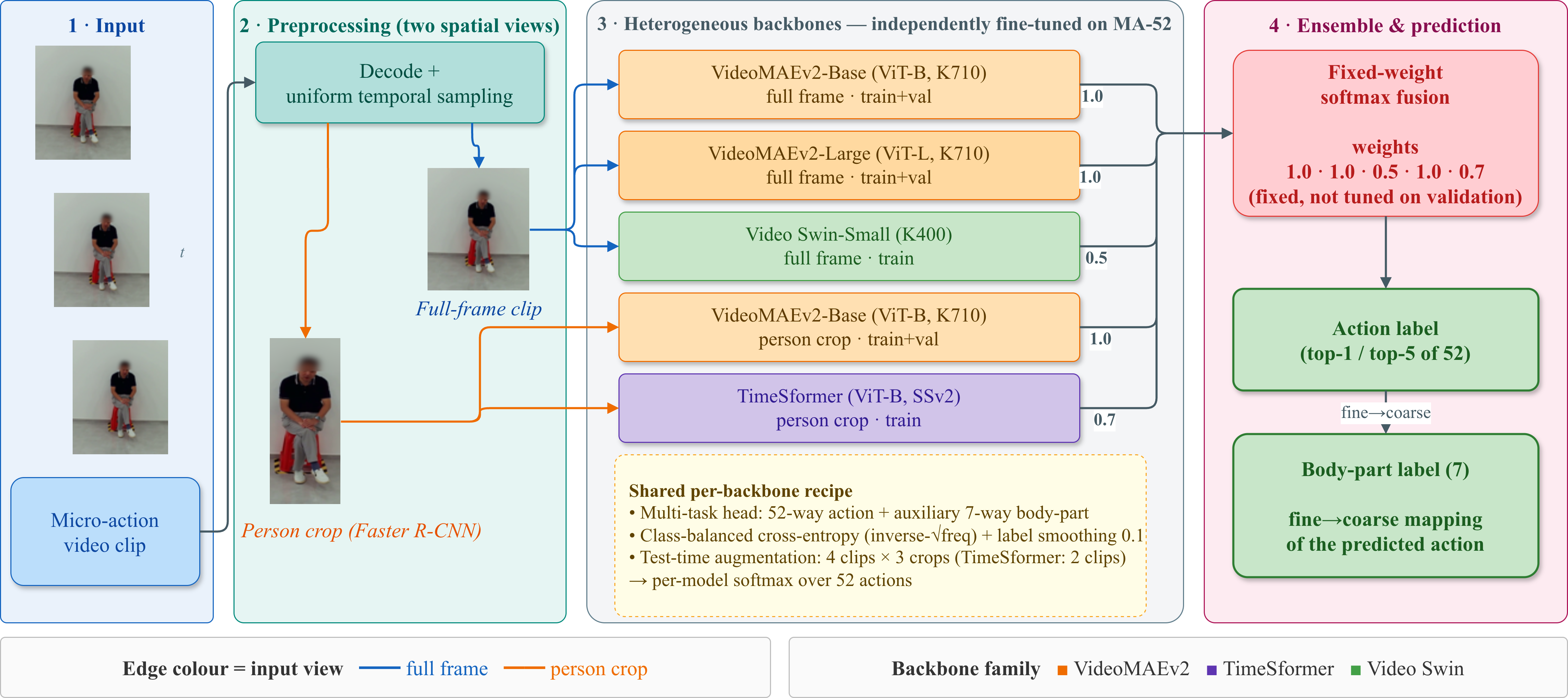}
\vspace{-1.5em}
\caption{Solution of team salaheiyo1 for Track 1.}
\vspace{-1.5em}
\label{fig:track1_top3}
\end{figure}

\ding{184}
Team ``salaheiyo1'' is from Aalto University.
As shown in Fig~\ref{fig:track1_top3}, this team adopts a heterogeneous multi-backbone ensemble framework that combines multiple complementary video backbones and person-centric views.
Their method uses uniform temporal sampling to construct full-frame clips, while Faster R-CNN~\cite{ren2015faster} is applied to generate person-cropped inputs. Multiple video backbones, including VideoMAEv2-Base~\cite{wang2023videomae}, VideoMAEv2-Large~\cite{wang2023videomae}, Video Swin-Small~\cite{liu2022video}, and TimeSformer~\cite{bertasius2021space}, are independently fine-tuned on these views to exploit diverse spatiotemporal representations. 
The team further improves performance with joint action-level and body-part-level classification, class-balanced training, and test-time augmentation. Finally, fixed-weight softmax fusion is used to ensemble the predictions from different models and views, yielding both action- and body-part labels.

\begin{figure}[t!]
\centering
\includegraphics[width=1\linewidth]{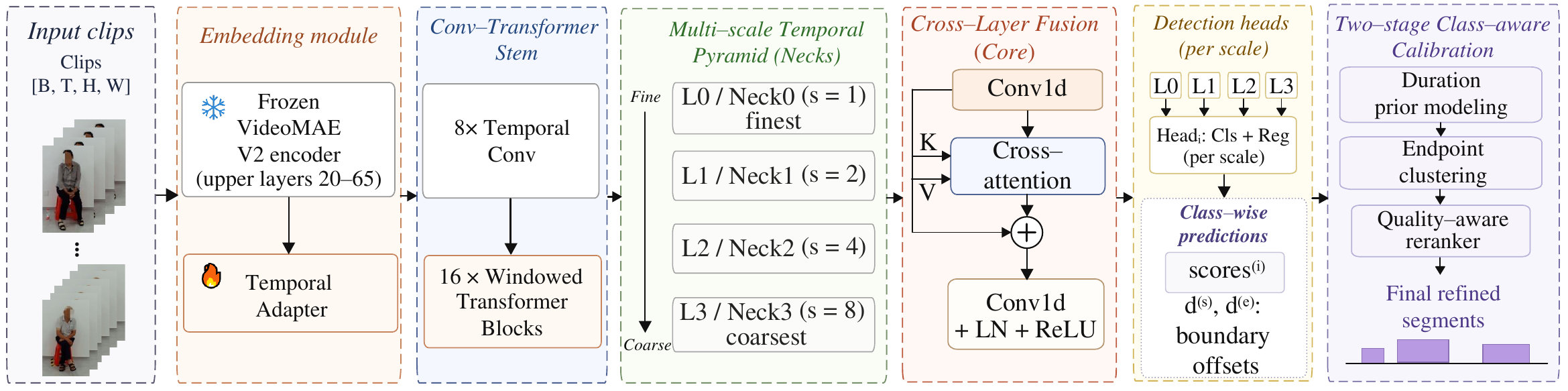}
\vspace{-1.5em}
\caption{Solution of team USTC-IAT-United1 for Track 2.}
\vspace{-1.5em}
\label{fig:track2_top1}
\end{figure}


\subsection{Multi-label Micro-Action Detection}

\ding{182} Team ``USTC-IAT-United1'' is from the University of Science and Technology of China. 
Inspired by the concise method description style in the MAC 2025 report~\cite{li2025mac}, their algorithm focuses on two key modules: temporal adapters and cross-layer fusion.
The first module is a temporal adapter inserted into a frozen VideoMAE-V2 backbone. Instead of fully fine-tuning the large video transformer, they freeze its original weights and train only lightweight adapter layers together with the detection head. 
The second module is a cross-layer fusion mechanism built on top of a multi-scale temporal feature pyramid. The detector keeps a full-resolution stem for short actions and then builds coarser temporal levels for longer context. 
To combine fine boundary detail and coarse semantic information, cross-layer fusion aligns features from different pyramid levels and exchanges information through cross-attention.

\begin{figure}[t]
\centering
\includegraphics[width=1\linewidth]{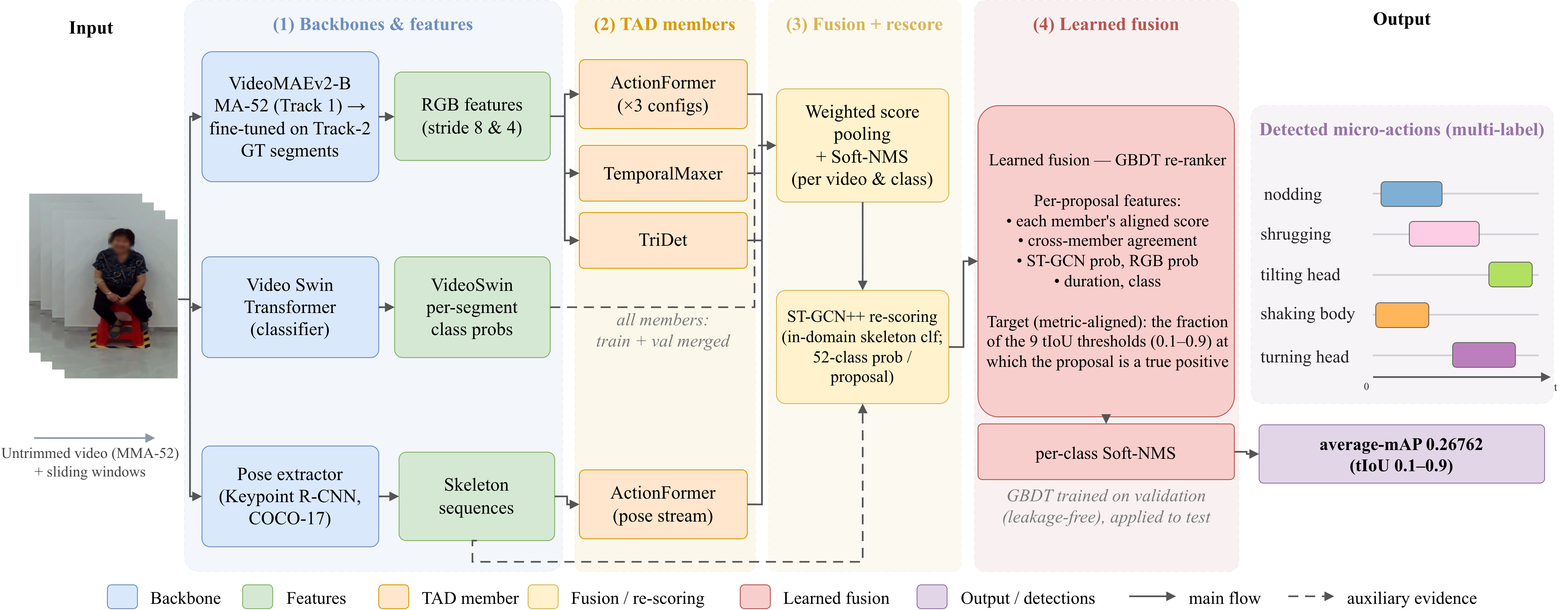}
\vspace{-1.5em}
\caption{Solution of team salaheiyo1 for Track 2.}
\vspace{-1.5em}
\label{fig:track2_top2}
\end{figure}
\ding{183} Team ``salaheiyo1'' is from Aalto University. They proposed a multi-backbone temporal-action-detection ensemble with a learned, metric-aligned fusion. As illustrated in Fig.\ref{fig:track2_top2}, the pipeline maps untrimmed videos to multi-label detections through four stages. A VideoMAEv2~\cite{wang2023videomae} encoder, first fine-tuned on the MA-52 micro-action recognition data, is further fine-tuned on Track-2's own ground-truth segments. A Video Swin Transformer~\cite{liu2022video} classifier and a 2-D human-pose extractor supply complementary appearance and skeleton streams. The members' proposals are combined per video and per class by weighted score pooling followed by Soft-NMS~\cite{bodla2017soft}. An in-domain ST-GCN++~\cite{duan2022pyskl} classifier then re-scores every surviving proposal, attaching a 52-class probability vector.

\begin{figure}[t]
\centering
\includegraphics[width=1.0\linewidth]{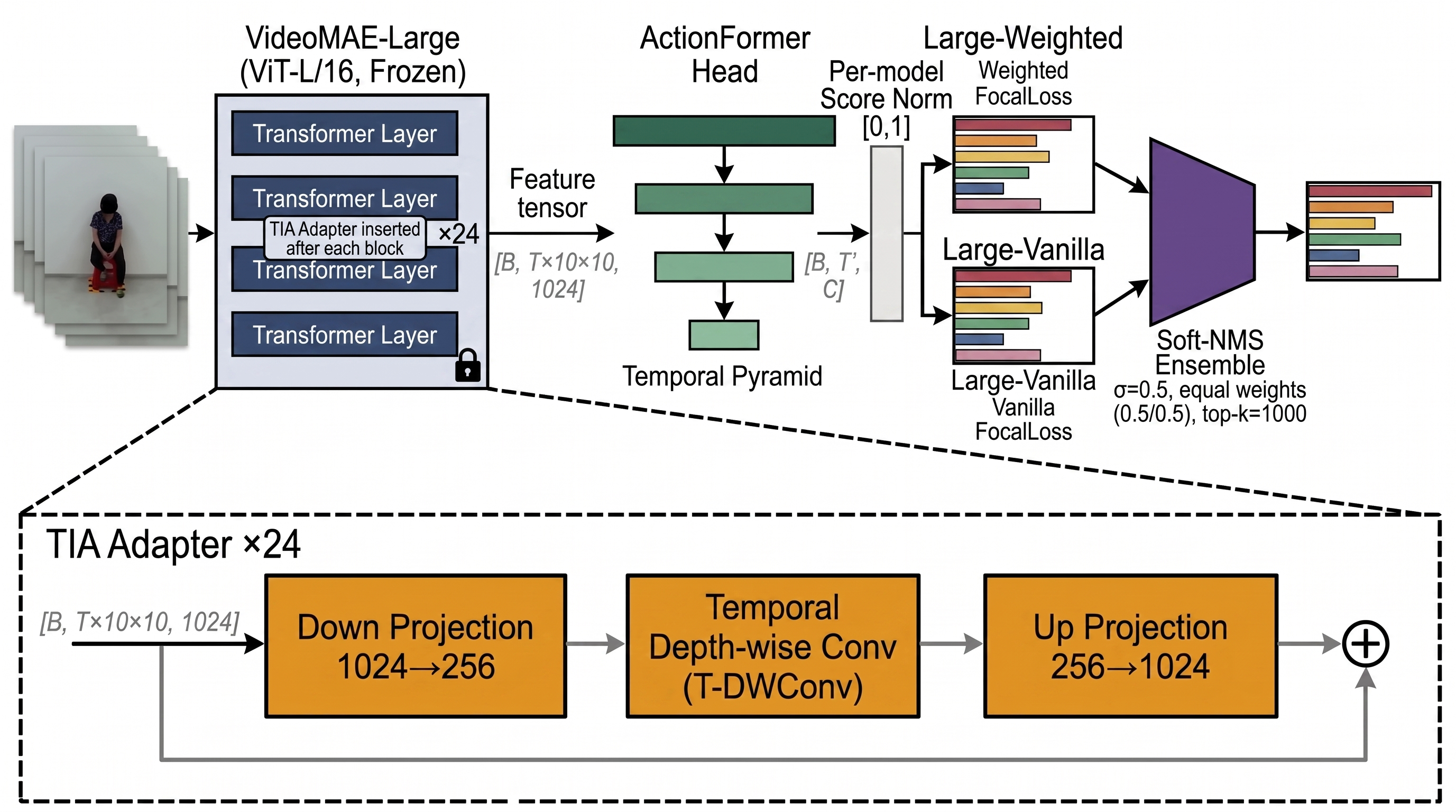}
\vspace{-1.5em}
\caption{Solution of team LR for Track 2.}
\vspace{-1.5em}
\label{fig:track2_top3}
\end{figure}

{\ding{184} Team ``LR''} is from Friedrich-Alexander-Universit\"{a}tErlangen-N\"{u}rnberg (FAU). 
The team built upon the AdaTAD~\cite{liu2024end} framework. Their solution is shown in Fig.~\ref{fig:track2_top3}. To detect micro-actions, the team adopted a frozen VideoMAE-Large backbone~\cite{wang2023videomae} combined with lightweight temporal adapters, enabling the model to learn task-specific representations without modifying the pre-trained weights. A Temporal-channel\sloppy Interaction Adapter bottleneck~\cite{liu2024end} was inserted after every transformer block. Each TIA adapter applied a down-projection, a temporal depth-wise convolution to aggregate context across adjacent frames, and an up-projection. An ActionFormer~\cite{zhang2022actionformer} detection head with a multi-scale temporal feature pyramid, Focal Loss classification, and DIoU regression produced the final temporal proposals. At inference, the two model variants were fused via a Gaussian Soft-NMS~\cite{bodla2017soft} ensemble.

\subsection{Fine-grained Micro-Action Understanding}

\begin{figure}[t]
\centering
\includegraphics[width=1.0\linewidth]{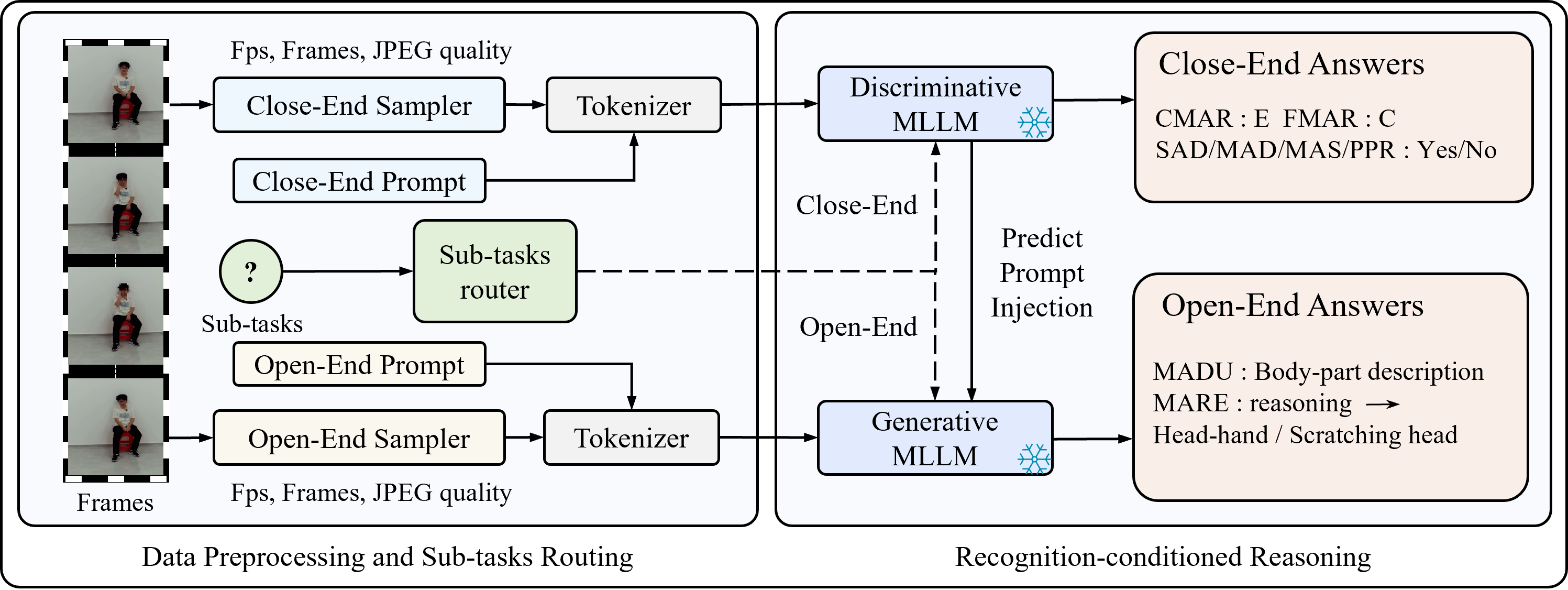}
\vspace{-1.5em}
\caption{Solution of team WHU-HUVPR for Track 3.}
\vspace{-1.5em}
\label{fig:track3_top1}
\end{figure}

\ding{182} Team ``WHU-HUVPR'' is from Wuhan University. As illustrated in Fig.~\ref{fig:track3_top1}, the team formulates Track 3 as a prompt-routing pipeline with recognition-conditioned reasoning. Closed-ended prompts are first used to produce compact recognition and relation judgments, and the predicted coarse- and fine-labels are then used as explicit priors for open-ended reasoning. This design decouples label selection from explanation generation: the recognition prompt provides a label anchor, while the reasoning prompt asks the model to justify it with visible motion evidence, body-part trajectory, temporal phase, and label-level distinctions. The method focuses on prompt orchestration and evidence-grounded explanation under a fixed-MLLM setting.

\begin{figure}[t]
\centering
\includegraphics[width=1.0\linewidth]{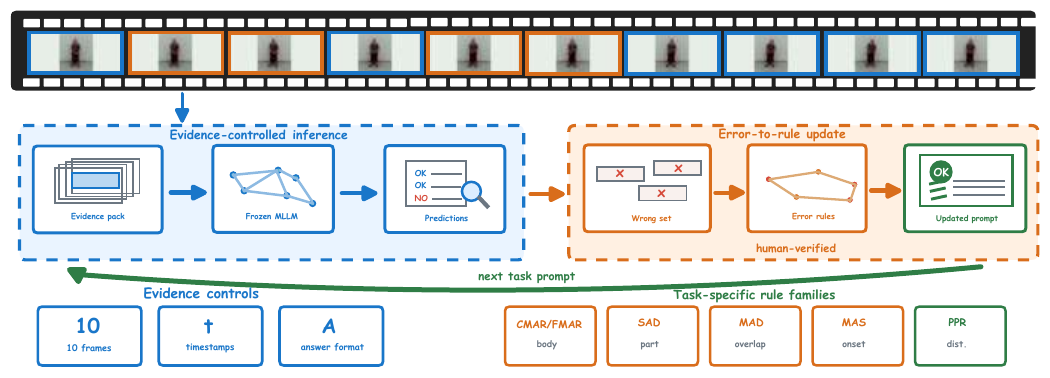}
\vspace{-1.5em}
\caption{Solution of the team GNC for Track 3.}
\vspace{-1.5em}
\label{fig:track3_top3}
\end{figure}

\ding{183} Team ``GNC'' is from Shanghai University. As shown in Fig.~\ref{fig:track3_top3}, the team proposes EDPO, an error-driven prompt optimization workflow for fixed MLLMs. The pipeline starts from baseline prompts, groups wrong predictions by sub-task, summarizes recurring failure types, and rewrites them into task-specific prompt rules, timestamped evidence checks, and constrained answer formats. 
Different rule groups are designed for different question types: CMAR and FMAR focus on refining category boundaries; SAD emphasizes weak local motion and phase cues; MAD handles overlapping body-part movements; MAS compares event onset order, and PPR verifies distance changes rather than static proximity. In this way, each prompt revision is traceable, as every added instruction corresponds to an observed error pattern.

\section{Conclusion and Future Directions}~\label{sec:conclusion}
As a continuing benchmark series, the 3rd MAC further advances micro-action analysis from recognition and detection toward fine-grained understanding. In the future, we plan to extend MAC toward emotion-aware micro-action understanding, encouraging models to infer affective states, social intentions, and psychological cues from subtle body behaviors. We also aim to develop more reliable evaluation protocols for assessing action grounding and reasoning consistency in multimodal large language models.

\begin{acks}
We sincerely thank the organizers of ACM MM 2026 for their support in hosting this challenge. 
We also thank the Data Chairs and the Program Committee for their coordination and continuous support. 
Finally, we extend our gratitude to all participating teams and individuals for their active involvement, valuable submissions, and continued interest in micro-action analysis. 
This work was supported by the National Natural Science Foundation of China (62272144, 72188101), the Anhui Provincial Natural Science Foundation (2408085J040), the National Key R\&D Program of China (2024YFB3311600), the Major Project of Anhui Provincial Science and Technology Breakthrough Program (202423k09020001), and the New Cornerstone Science Foundation through the XPLORER PRIZE. 
\end{acks}

\bibliographystyle{ACM-Reference-Format}
\balance
\bibliography{sample-base}

\end{document}